\pdfoutput=1

\documentclass[11pt]{article}

\usepackage{acl}

\usepackage{times}
\usepackage{latexsym}

\usepackage[T1]{fontenc}

\usepackage[utf8]{inputenc}

\usepackage{microtype}

\usepackage{inconsolata}

\usepackage{graphicx}

\usepackage{booktabs}
\usepackage{amsmath}
\usepackage{multicol}
\usepackage{tabularx}

\usepackage[framemethod=TikZ]{mdframed}
\usepackage{listings}
\newmdenv[%
    backgroundcolor=gray!10,
    linecolor=black,
    outerlinewidth=0.5pt,
    roundcorner=1mm,
    skipabove=\topsep,
    skipbelow=\topsep,
    font=\ttfamily\footnotesize,
]{promptbox}

\newcommand{\aspace}{\hspace{0.5em}}

\title{Exploring Pretraining via Active Forgetting for Improving Cross Lingual Transfer for Decoder Language Models}

\author{First Author \\
  Affiliation / Address line 1 \\
  Affiliation / Address line 2 \\
  Affiliation / Address line 3 \\
  \texttt{email@domain} \\\And
  Second Author \\
  Affiliation / Address line 1 \\
  Affiliation / Address line 2 \\
  Affiliation / Address line 3 \\
  \texttt{email@domain} \\}

\author{
Divyanshu Aggarwal\thanks{\enspace Equal contribution} \aspace Ashutosh Sathe\thanks{\enspace Work done during internship at MSRI}\footnotemark[1] \aspace Sunayana Sitaram\\
Microsoft Research India \\
\small{\texttt{t-daggarwal@microsoft.com, absathe@cse.iitb.ac.in, sunayana.sitaram@microsoft.com}} \\
}

\begin{document}
\maketitle
\begin{abstract}
Large Language Models (LLMs) demonstrate exceptional capabilities in a multitude of NLP tasks. However, the efficacy of such models to languages other than English is often limited. Prior works have shown that encoder-only models such as BERT or XLM-RoBERTa show impressive cross lingual transfer of their capabilities from English to other languages. In this work, we propose a pretraining strategy that uses active forgetting to achieve similar cross lingual transfer in decoder-only LLMs. We show that LLMs pretrained with active forgetting are highly effective when adapting to new and unseen languages. Through extensive experimentation, we find that LLMs pretrained with active forgetting are able to learn better multilingual representations which translates to better performance in many downstream tasks.
\end{abstract}

\section{Introduction}

Despite demonstrating excellent performance on English, LLM performance on multilingual benchmarks is often limited \citep{ahuja-etal-2023-mega, ahuja2024megaverse}. A common method to introduce new languages to existing LLMs involves vocabulary expansion and retraining token embeddings \citep{balachandran2023tamilllama, cui2024efficient}. In many cases, these models are further finetuned on translations of English instruction tuning datasets such as \citet{li2023bactrianx, wei2023polylm, singh2024aya}. Building multilingual LLMs by simply having a large number of languages in the pretraining also does not work well due to the so-called ``curse of multilinguality'' \citep{conneau2020unsupervisedcrosslingualrepresentationlearning}. Interestingly, encoder-only LLMs demonstrate cross lingual transfer -- a phenomenon where the LLMs improve performance on non-English languages despite being trained only on English data. Past work has worked on improving such cross-lingual transfer \citep{pfeiffer2020madx, parovic2023crosslingual, chen2024improving} in encoder-only LLMs such as BERT or XLM-RoBERTa but less attention has been paid to cross-lingual transfer of decoder-only autoregressive LLMs.



\begin{figure}[t]
    \centering
    \includegraphics[width=0.5\textwidth]{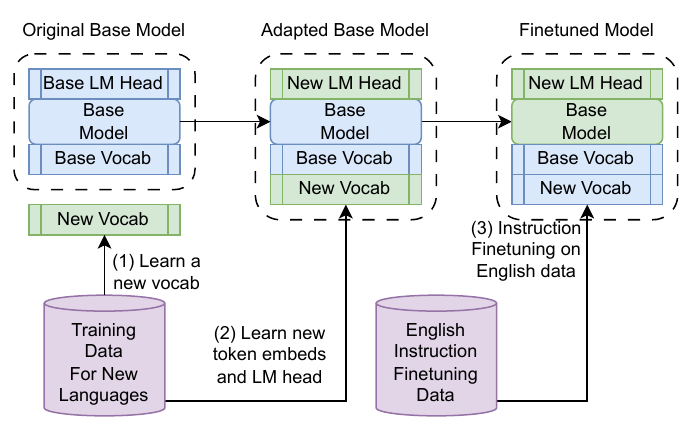}
    \caption{\textbf{Adapting base LLMs to new languages.} We show that if the base LLM is pretrained using active forgetting, it improves cross lingual transfer capabilities of the resultant English-only instruction finetuned model.}
    \label{fig:adaptign_procedure}
\end{figure}

We find (Sec. \ref{sec:discussions}) that the common method (depicted in Figure \ref{fig:adaptign_procedure}) to adapt base LLMs to a narrow set of target languages via vocabulary expansion results in improvements only to the target languages. Our results show that such adaptation significantly worsens the overall multilingual capabilities of the resultant model thereby limiting the cross lingual transfer capabilities of LLMs. Methods like zero-shot tokenizer transfer \citep{minixhofer2024zeroshot} show promising results when adapting (decoder-only) base LLMs to new tasks such as programming \textit{without} vocabulary expansion, but their applicability in the multilingual context has not been studied well. On the other hand, \citep{chen2024improving} show that training encoder-only LLMs with active forgetting leads to better language ``plasticity'' and further improves cross lingual transfer.



In this work, we want to improve the cross lingual transfer abilities of autoregressive LLMs by pretraining them using active forgetting, proposed by \citet{chen2024improving}. We show that models pretrained via active forgetting are better at adapting to new languages with lower degradation in performance on other languages. We also show that models pretrained with active forgetting have better perplexity and isotropy as compared to vanilla pretrained and adapted large language models. Our contributions are listed below: 
\begin{enumerate}
    \item We find that the common method of adapting base LLMs to newer languages (Figure \ref{fig:adaptign_procedure}) leads to improvements in performance \textit{only} on the newer languages at the cost of performance of other languages.
    \item We illustrate that base LLMs pretrained with active forgetting lead to higher quality multilingual representations.
    \item These improved representations also lead to better cross lingual transfer. Active forgetting based LLMs outperform the baselines on 6 out of 7 multilingual benchmarks. 
    
\end{enumerate}



\section{Related Work}
\label{sec:related work}
\paragraph{Multilingual Language Modelling}
Efforts like BLOOM \cite{workshop2023bloom} and PolyLM \cite{wei2023polylm} have created large multilingual LLMs, however, instruction finetuned LLMs are more desirable over pretrained models in real world usecases due to their zero shot instruction following capabilities. Multilingual instruction tuning datasets have been introduced by \citet{wei2023polylm}, \citet{li2023bactrianx} and \citet{singh2024aya} to further enhance the instruction following capabilities of these models in multilingual settings. Most of these datasets are synthetically generated except Aya which is human curated. The costs of high quality human annotations show the need for models with high cross lingual transfer capabilities.


\paragraph{Cross Lingual Transfer}
Earlier works have shown that crosslingual transfer can be beneficial for multilingual pretrained models to gain task ability from English-only labelled data \cite{rajaee-monz-2024-analyzing, deb-etal-2023-zero, parovic2023crosslingual, zhao2024adamergex}. While these techniques are effective to a certain extent, they are limited by the multilingual abilities of the pretrained model. Moreover, they cannot be effective for the low resource settings even on multilingual models due to poorer representation of rarer tokens in pretraining corpus and high token fertility for morphologically richer languages.

\paragraph{Language Adaptation}
Creating language models that can learn newer languages successively without further pretraining or by training minimal additional parameters from unlabelled data is of significant interest to reearch community \cite{chen2024improving, pfeiffer2020madx, zhao2024llama}. While these techniques improve the language capabilities on the newer languages, our results show that this improvement comes at the cost of performance on other languages. We focus on improving the pretraining of the base LLM itself with the intention of improving performance of such language adaptation techniques post training.

\section{Method}
\label{sec:method}

\citet{chern-etal-2023-improving} propose using ``active forgetting'' based pretraining where token embeddings of the model are reset to random embeddings after every $k$ steps of pretraining. They find that using active forgetting to pretrain encoder-only models improves their cross lingual transfer i.e. finetuning only on task-specific English labelled data improves task performance on non-English languages as well. In this work, we study benefits of active forgetting to train and adapt decoder-only models to new and unseen languages. Figure \ref{fig:adaptign_procedure} shows standard procedure of introducing new languages to the base LLM through vocabulary expansion. 

Specifically, we are given a base LLM $\mathcal{M}_\text{base}$ with vocabulary $\mathcal{V}$ which we wish to adapt to $L$ new languages. We assume access to a reasonably sized corpus $\mathcal{D}_\text{train}^L$ consisting of unstructured text of $L$ languages. In the adaptation process, first a new vocabulary $\mathcal{V}^L$ is learned over $\mathcal{D}_\text{train}^L$ and merged with $\mathcal{V}$ to form a larger vocabulary $\mathcal{V}_\text{merged}$. In the second stage, the language modeling head of $\mathcal{M}$ is replaced to be of the appropriate size i.e. $|\mathcal{V}_\text{merged}|$. Then the new language modeling head and token embeddings of newly added tokens (i.e. $\mathcal{V}_\text{merged} - \mathcal{V}$) are learned with standard language modeling training over $\mathcal{D}_\text{train}^L$. Notice that the entire Transformer stack of $\mathcal{M}$ and token embeddings of $\mathcal{V}$ are held frozen during this training. The resultant model at the end of second stage is denoted by $\mathcal{M}_\text{adapted}$ and has language modeling head of the size $|\mathcal{V}_\text{merged}|$. In the final stage, we instruction finetune the $\mathcal{M}_\text{adapted}$ on English only data to get $\mathcal{M}_\text{adapted}^\text{finetuned}$ which is evaluated on multilingual benchmarks to assess its cross lingual transfer capoabilities. Our hypothesis is that if $\mathcal{M}_\text{base}$ is pretrained using active forgetting, the corresponding $\mathcal{M}_\text{adapted}^\text{finetuned}$ will be better at cross lingual transfer.

\section{Experiments}
\label{sec:experiments}
\paragraph{Training Setup} We pretrain our $\mathcal{M}_\text{base}$ on Wikipedia dumps\footnote{\tiny{\url{https://huggingface.co/datasets/wikimedia/wikipedia} 20231101 dump}} of 12 languages (referred as ``pretraining'' languages) from \citet{pinchofmulti}.  The adaptation dataset $\mathcal{D}_\text{train}^L$ consists of Wikipedia dumps of 14 new languages (referred as ``adapting'' languages) disjoint from the pretraining languages. The exact languages are presented in Table \ref{tab:language_details}. In our results, ``BA'' refers to ``Baseline Adapted'' i.e. $\mathcal{M}_\text{adapted}^\text{finetuned}$ where $\mathcal{M}_\text{base}$ was trained with standard optimization. ``AFA'' refers to ``Active Forgetting Adapted'' i.e. $\mathcal{M}_\text{adapted}^\text{finetuned}$ where $\mathcal{M}_\text{base}$ was trained with active forgetting. We also present results on ``Baseline'' which refers to $\mathcal{M}_\text{base}^\text{finetuned}$ i.e. instruction tuned $\mathcal{M}_\text{base}$ without adaptation. We experiment with $\mathcal{M}_\text{base}$ of 3 different sizes and use OpenOrca \citep{OpenOrca} as our English instruction tuning dataset and contains 2.91M data points.

\paragraph{Evaluation Setup} We follow \citet{aggarwal2024maple} to evaluate multilingual capabilities of our models using 6 multilingual benchmarks. Additionally, we establish superiority of the active forgetting pretrained models by measuring isotropy of their embeddings \citep{ethayarajh-2019-contextual} and model perplexity on 50 languages (26 new languages not in ``pretraining'' or ``adapting'' as shown in Table \ref{tab:language_details}) in mC4 \citep{xue-etal-2021-mt5}. We also evaluate the 4-shot translation (English-to-X) performance of the models to the same set of 50 languages using the FLORES-200 dataset \citep{nllb-22}.

\section{Discussion}
\label{sec:discussions}

\paragraph{Active Forgetting Leads to Better Language Adaptation}

\begin{table}[tb!]
\small
\centering
\begin{tabular}{lrrrr}
\toprule
Model & $\mu_\text{pretraining}$  & $\mu_\text{adapting}$ & $\mu_\text{other}$  & $\mu_\text{overall}$ \\
\midrule
\multicolumn{5}{c}{Number of parameters = 400M}\\
\midrule
Baseline & \bfseries 25.041 & 31.440 & 34.663 & 31.451\\
BA & 25.097 & 31.405 & 36.993 & 32.573\\
AFA & 25.180 & \bfseries 30.373 & \bfseries 34.345 & \bfseries 31.033\\
\midrule
\multicolumn{5}{c}{Number of parameters = 782M}\\
\midrule
Baseline & 22.826 & 29.382 & 31.099 & 28.633\\
BA & 23.155 & 28.924 & 33.766 & 29.864\\
AFA & \bfseries 22.727 & \bfseries 27.949 & \bfseries 31.047 & \bfseries 28.183\\
\midrule
\multicolumn{5}{c}{Number of parameters = 1.6B}\\
\midrule
Baseline & 20.745 & 26.831 & 28.497 & 26.170\\
BA & 20.828 & 26.117 & 30.616 & 27.007\\
AFA & \bfseries 20.654 & \bfseries 25.048 & \bfseries 28.386 & \bfseries 25.596\\
\midrule
\multicolumn{5}{c}{Number of parameters = 2.8B}\\
\midrule
Baseline & 20.887 & 26.345 & 28.689 & 26.198\\
BA & 20.958 & 25.969 & 30.768 & 27.034\\
AFA & \bfseries 20.716 & \bfseries 24.858 & \bfseries 28.395 & \bfseries 25.621\\
\bottomrule
\end{tabular}

\caption{Detailed results on perplexity (Lower is Better). BA refers to Baseline adapted model. AFA refers to Active Forgetting adapted model. $\mu_\text{pretraining}$ refers to performance averaged over languages in the pretraining split. $\mu_\text{adapting}$ refers to averaging over languages in the adapting split. $\mu_\text{other}$ refers to averaging on languages that are in neither split. $\mu_\text{overall}$ refers to the average over all languages.
}
\label{tab:short_perplexity}
\end{table}
\begin{table}[t!]
\small
\centering
\begin{tabular}{lrrrr}
\toprule
Model & $\mu_\text{pretraining}$ & $\mu_\text{adapting}$  & $\mu_\text{other}$ & $\mu_\text{overall}$  \\
\midrule
\multicolumn{5}{c}{Number of parameters = 400M}\\
\midrule
Baseline & 0.683 & 0.663 & 0.667 & 0.670\\
BA & 0.659 & 0.651 & 0.678 & 0.666\\
AFA & \bfseries 0.640 & \bfseries 0.624 & \bfseries 0.640 & \bfseries 0.636\\
\midrule
\multicolumn{5}{c}{Number of parameters = 782M}\\
\midrule
Baseline & 0.610 & 0.607 & 0.612 & 0.610\\
BA & 0.602 & 0.593 & 0.618 & 0.607\\
AFA & \bfseries 0.587 & \bfseries 0.566 & \bfseries 0.588 & \bfseries 0.582\\
\midrule
\multicolumn{5}{c}{Number of parameters = 1.6B}\\
\midrule
Baseline & 0.549 & 0.550 & 0.562 & 0.555\\
BA & 0.555 & 0.548 & 0.560 & 0.555\\
AFA & \bfseries 0.531 & \bfseries 0.513 & \bfseries 0.530 & \bfseries 0.525\\
\midrule
\multicolumn{5}{c}{Number of parameters = 2.8B}\\
\midrule
Baseline & \bfseries 0.504 & \bfseries 0.506 & \bfseries  0.506 & \bfseries 0.505\\
BA & 0.506 & 0.508 & \bfseries 0.506 & 0.507\\
AFA & \bfseries 0.504 & \bfseries 0.506 & \bfseries 0.506 & \bfseries 0.505\\
\bottomrule
\end{tabular}
\caption{Detailed results on isotropy (Lower is Better). All the abbrevations are same as in Table \ref{tab:short_perplexity}.}
\label{tab:short_isotropy}
\end{table}

We study the intrinsic properties of the adapted models $\mathcal{M}_\text{adapted}$ using perplexity on mC4 and isotropy i.e. self similarity of contextual embeddings \citep{ethayarajh-2019-contextual}. As shown in Table \ref{tab:short_perplexity}, we find that AFA models achieve consistently lower perplexity than both ``Baseline'' and ``BA''. Moreover, AFA models are also able to better contextualize a sentence over all languages as observed by lower self similarity (isotropy) scores in Table \ref{tab:short_isotropy}. This suggests that the quality of multilingual representations of AFA is better than other models.

\paragraph{Active Forgetting Improves Cross-Lingual Transfer}

\begin{figure}[t!]
    \centering
    \includegraphics[width=0.5\textwidth]{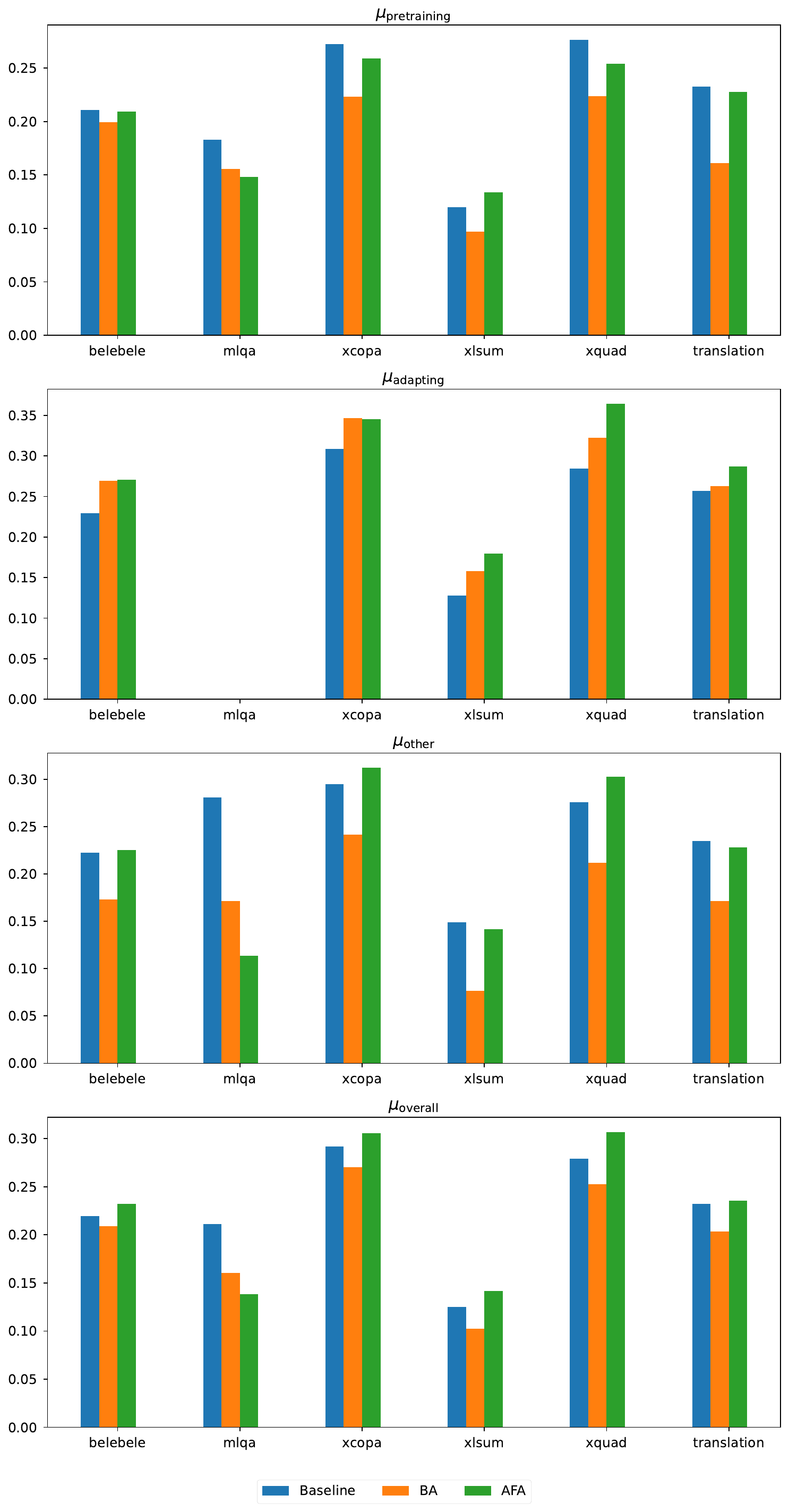}
    \caption{Task wise performance comparison for the 2.8 billion parameter models. Higher is better for all tasks.
    }
    \label{fig:taskwise_2.8b}
\end{figure}

In Figure \ref{fig:taskwise_1.6b}, we compare performance of our models on various multilingual benchmarks similar to \citet{aggarwal2024maple} and the translation task. We find that despite instruction tuning only on English, AFA models show improvements across all language classes. AFA outperforms both Baseline and BA models on 6 out of 7 tasks in our evaluation suite. More importantly, we find that BA models often are worse overall ($\mu_\text{overall}$) as compared to Baseline. This reaffirms findings by \citet{pinchofmulti} where if the base model is already multilingual, adapting to a narrow set of languages can \textit{worsen} the overall performance. AFA models on the other hand do not seem to suffer from the same limitation.


\paragraph{Analysis on Language Class and Model Size}

\begin{table}[t]
\small
\centering
\begin{tabular}{lrrrr}
\toprule
Model & $\mu_\text{pretraining}$ & $\mu_\text{adapting}$ & $\mu_\text{other}$ & $\mu_\text{overall}$ \\
\midrule
\multicolumn{5}{c}{Number of parameters = 400M}\\
\midrule
Baseline & 0.080 & 0.084 & \bfseries 0.100 & 0.091\\
BA & \bfseries 0.101 & 0.092 & 0.074 & 0.086\\
AFA & 0.078 & \bfseries 0.103 & 0.098 & \bfseries 0.094\\
\midrule
\multicolumn{5}{c}{Number of parameters = 782M}\\
\midrule
Baseline & \bfseries 0.162 & 0.138 & 0.154 & 0.152\\
BA & 0.127 & \bfseries 0.208 & 0.119 & 0.146\\
AFA & 0.158 & 0.190 & \bfseries 0.180 & \bfseries 0.178\\
\midrule
\multicolumn{5}{c}{Number of parameters = 1.6B}\\
\midrule
Baseline & \bfseries 0.208 & 0.197 & \bfseries 0.221 & 0.211\\
BA & 0.134 & \bfseries 0.274 & 0.147 & 0.180\\
AFA & 0.202 & 0.254 & 0.198 & \bfseries 0.215\\
\midrule
\multicolumn{5}{c}{Number of parameters = 2.8B}\\
\midrule
Baseline & \bfseries 0.241 & 0.255 & \bfseries 0.245 & 0.237\\
BA & 0.163 & 0.255 & 0.174 & 0.205\\
AFA & 0.240 & \textbf{0.288} & 0.229 & \bfseries 0.239\\
\bottomrule
\end{tabular}
\caption{Detailed results on translation (en-to-XX) on the subset of languages from FLORES-200 \citep{nllb-22}. All the abbrevations are same as in Table \ref{tab:short_perplexity} (Higher is better). We use BLEU Score as the metric \cite{papineni-etal-2002-bleu}. 
}
\label{tab:short_translation}
\end{table}

\begin{figure}[thbp!]
    \centering
    \includegraphics[scale=0.85]{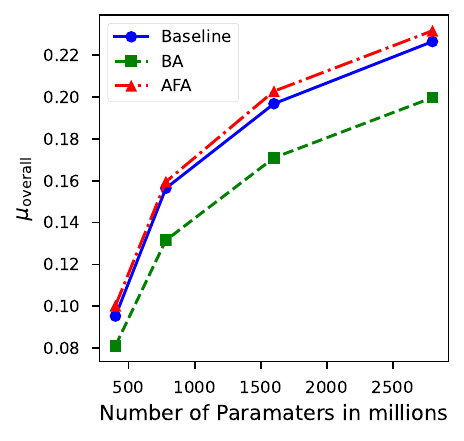}
    \caption{\textbf{Effect of model scale.} Average task performance of models against the model parameter count. We consider the same tasks in figure \ref{fig:taskwise_2.8b}}
    \label{fig:model_scale}
\end{figure}

We study how language adaptation affects performance on each language class (``pretraining'', ``adapting'' or ``other'')  by studying performance on translation in Table \ref{tab:short_translation}. We find that BA models show significantly better performance (as compared to Baseline) on ``adapting'' languages at all model scales. Moreover, improvement of BA over Baseline is larger than improvement of AFA over Baseline for larger models and BA models seem to degrade performance on all other language classes leading to worse overall($\mu_\text{overall}$) scores as observed in Figure \ref{fig:model_scale}.

\section{Conclusion and Future Work}
\label{sec:conclusion}

In this work, we show that pretraining with active forgetting can improve language adaptability and cross lingual transfer capabilities of autoregressive (decoder-only) language models. In our experiments, we found that a base LLM that is pretrained with active forgetting and instruction tuned only with English data leads to improvements across all languages in 6 out of 7 multilingual benchmarks. We observed this behavior to be consistent at all the model sizes we tried. The improvements in these downstream tasks can be attributed to the active forgetting models learning better multilingual representations. We hope that these findings encourage pretraining of larger LLMs with active forgetting. Future work can also explored effective language adaptation methods to adapt an \textit{existing} finetuned LLM to new languages.

\section*{Limitations}

The methods described in this work are aimed at improving \textit{pretraining} of multilingual LLMs. As such, these cannot be directly applied to existing LLMs. An interesting direction to explore could be to take intermediate checkpoints of open source LLMs such as TinyLlama \citep{zhang2024tinyllama} or OLMo \citep{groeneveld2024olmo} and simulate active forgetting by resetting their embeddings then continuing to train on $\mathcal{D}_\text{train}^L$. Finally, our training and evaluation suite consisted primarily of training language models of size that could comfortably fit in our compute budget. Moreover, the data is used (10 billion tokens in total) is much lesser than models like XLM-R which were trained on much larger data with more than 100 billion in much more languages, which can give state of the art performance on our evaluation with much lesser model parameter size, since the data is larger, the compute FLoPs used to train XLM-R is much larger than what we used despite our models being larger.  Further experiments and evaluation is needed to study efficacy of pretraining with active forgetting on larger scale models.

\section*{Ethics Statement}

The proposed method of pretraining directly affects the token embeddings of an LLM. While we find that these lead to better representations in terms of multilinguality, special care must be taken before deploying such LLMs. A thorough study of their overall capabilities as well as intrinsic and extrinsic biases must be performed before deploying such LLMs to any public facing interface.

\bibliography{custom,anthology}

\clearpage

\appendix

\section{Details on Computational Resources}

\begin{table}
\small
\centering
\begin{tabularx}{0.45\textwidth}{lX}
\toprule
\textbf{Hyperparameter}    & \textbf{Value} \\ \midrule
Learning rate              & $1\times 10^{-4}$ \\
Number of steps            & $150,000$ \\ 
Global batch size          & $128$ \\
Block size                 & $4096$ \\
Scheduler                  & Cosine \\
Warmup                     & Linear \\
Warmup steps               & 10\% \\
Optimizer                  & AdamW \tiny{\citep{loshchilov2018decoupled}}\\
Weight decay               & $0$              \\
Embed. reset steps         & $10,000$ \\
\bottomrule
\end{tabularx}
\caption{Hyperparameters for pretraining. Only the active forgetting models reset their token embeddings every ``Embed. reset steps''. Note that embeddings are \textit{not} reset after the final step.}
\label{tab:indiv_pt}
\end{table}

\begin{table}
\small
\centering
\begin{tabularx}{0.45\textwidth}{lX}
\toprule
\textbf{Hyperparameter}    & \textbf{Value} \\ \midrule
Learning rate              & $1\times 10^{-6}$ \\
Epochs                     & $5$ \\ 
Global batch size          & $16$ \\
Scheduler                  & Cosine \\
Warmup                     & Linear \\
Warmup steps               & 10\% \\
Optimizer                  & AdamW \tiny{\citep{loshchilov2018decoupled}}\\
Weight decay               & $0$              \\
\bottomrule
\end{tabularx}
\caption{Hyperparameters for finetuning.}
\label{tab:indiv_ft}
\end{table}

Our base LLM uses Mistral architecture downscaled to fit our compute resources. We explore 3 configurations with total parameter counts of 400M, 782M and 1.6B respectively by reducing the hidden dimensions, number of attention heads and the total number of Transformer blocks. The vocabulary size of the base model was kept fixedd at 32000 while the adapting vocabulary was allowed to merge 16000 more tokens leading to $|\mathcal{V}_\text{merged}| = 48000$. All our experiments are run on a single NVIDIA A100 GPU with 80 GB VRAM. The total GPU hours for all experiments and evaluations come out to roughly 650 hours. The hyperparameters for pretraining and finetuning runs are presented in Table \ref{tab:indiv_pt} and Table \ref{tab:indiv_ft} respectively.

\section{Details on Language Classes}
\label{sec:lang-details}
\begin{table}[]
    \centering
    \begin{tabular}{lp{4.5cm}}
    \toprule
        group & languages \\
    \midrule
        pretraining (12) & ar, zh, cs, en, et, fi, he, hi, it, ru, es, sw \\
        \midrule
        adapting (14) & ja, fr, pt, nl, se, tr, da, no, ko, pl, hu, th, mr, gu \\
        \midrule
        other (26) & af, bg, bn, de, id, ml, sv, ta, te, ur, vi, tg, ka, sq, ps, sr, az, my, co, iw, mn, st, sk, ha \\
    \bottomrule
    \end{tabular}
    \caption{Language Details}
    \label{tab:language_details}

\end{table}

In Table \ref{tab:language_details}, we present the 2 main classes of languages relevant to our experiments. Specifically, all base models (with or without active forgetting) are pretrained on the ``pretraining'' languages. Each base model is then adapted to the 14 ``adapting'' languages together.  Any language that is not in either of these 2 classes is considered ``other''. The ``other'' in Table \ref{tab:language_details} refers to the additional languages we used for perplexity and isotropy analysis.

\section{Evaluation Setting and Prompts}
\label{sec:eval_prompts}

We use lm-evaluation-harness \footnote{\url{https://github.com/EleutherAI/lm-evaluation-harness}} for our evaluation experiments with default settings. 

The evaluation prompts for all the tasks in our evaluation suite (6 tasks from \citet{aggarwal2024maple} and Translation on FLORES-200) are presented in Figure \ref{fig:XNLI prompt} to Figure \ref{fig:Tanslation prompt}.






\begin{figure}[!h]
\centering
\begin{promptbox}
<|im\_start|>system\\
You are a large language model trained to solve multiple NLP tasks accurately. For any given NLP task, you must produce an output that is factually correct and succinct.\\
<|im\_end|>\\
<|im\_start|>user\\

The task is to perform open-domain commonsense causal reasoning. You will be provided a premise and two alternatives, where the task is to select the alternative that more plausibly has a causal relation with the premise. Answer as concisely as possible in the same format as the examples below:
\\

Given this premise: 
\\
\{\{premise\}\}
\\ 
What's the best option?
\\
-choice1 : \{\{choice1\}\}
\\
-choice2 : \{\{choice2\}\}
\\
We are looking for \{\% if question == "cause" \%\} a cause \{\% else \%\} an effect \{\% endif \%\}
\\ 
<|im\_end|>\\
<|im\_start|>assistant
\\
\end{promptbox}
\caption{XCOPA Prompt}
\label{fig:XCOPA prompt}
\end{figure}


\begin{figure}[!h]
\centering
\begin{promptbox}
<|im\_start|>system\\
You are a large language model trained to solve multiple NLP tasks accurately. For any given NLP task, you must produce an output that is factually correct and succinct.\\
<|im\_end|>\\
<|im\_start|>user\\

The task is to perform reading comprehension task. Given the following passage, query, and answer choices, output the letter corresponding to the correct answer.
\\ \\
Passage: \{\{flores\_passage\}\}
\\
Query: \{\{question\}\}
\\
Choices:\\
A: \{\{mc\_answer1\}\}
\\
B: \{\{mc\_answer2\}\}
\\
C: \{\{mc\_answer3\}\}
\\
D: \{\{mc\_answer4\}\}
\\ \\
<|im\_end|>\\
<|im\_start|>assistant
\\
\end{promptbox}
\caption{Belebele Prompt}
\label{fig:Belebele prompt}
\end{figure}


\begin{figure}[!h]
\centering
\begin{promptbox}
<|im\_start|>system\\
You are a large language model trained to solve multiple NLP tasks accurately. For any given NLP task, you must produce an output that is factually correct and succinct.\\
<|im\_end|>\\
<|im\_start|>user\\

The task is to solve reading comprehension problems. You will be provided questions on a set of passages and you will need to provide the answer as it appears in the passage. The answer should be in the same language as the question and the passage.
\\ \\
Context:\{\{context\}\}
\\
Question:\{\{question\}\}
\\ \\
Referring to the passage above, the correct answer to the given question is
\\ \\
<|im\_end|>\\
<|im\_start|>assistant
\\
\end{promptbox}
\caption{MLQA Prompt}
\label{fig:MLQA prompt}
\end{figure}


\begin{figure}[!h]
\centering
\begin{promptbox}
<|im\_start|>system\\
You are a large language model trained to solve multiple NLP tasks accurately. For any given NLP task, you must produce an output that is factually correct and succinct.\\
<|im\_end|>\\
<|im\_start|>user\\

The task is to solve reading comprehension problems. You will be provided questions on a set of passages and you will need to provide the answer as it appears in the passage. The answer should be in the same language as the question and the passage.
\\ \\
Context:\{\{context\}\}
\\
Question:\{\{question\}\}
\\ \\
Referring to the passage above, the correct answer to the given question is
\\ \\
<|im\_end|>\\
<|im\_start|>assistant
\\
\end{promptbox}
\caption{XQUAD Prompt}
\label{fig:XQUAD prompt}
\end{figure}


\begin{figure}[!h]
\centering
\begin{promptbox}
<|im\_start|>system\\
You are a large language model trained to solve multiple NLP tasks accurately. For any given NLP task, you must produce an output that is factually correct and succinct.\\
<|im\_end|>\\
<|im\_start|>user\\

The task is to summarize any given article. You should summarize all important information concisely in the same language in which you have been provided the document. Following the examples provided below:
\\ \\
\{\{text\}\}
\\ \\
<|im\_end|>\\
<|im\_start|>assistant
\\
\end{promptbox}
\caption{XLSUM Prompt}
\label{fig:XLSUM prompt}
\end{figure}


\begin{figure}[!h]
\centering
\begin{promptbox}
<|im\_start|>system\\
You are a large language model trained to solve multiple NLP tasks accurately. For any given NLP task, you must produce an output that is factually correct and succinct.\\
<|im\_end|>\\
<|im\_start|>user\\

The task is to translate the given sentence in English to language \{\{language\}\}. There are 4 examples provided below. Produce the translation of the 5th sentence:
\\ \\
\{\{text\}\}
\\ \\
<|im\_end|>\\
<|im\_start|>assistant
\\
\end{promptbox}
\caption{Translation Prompt}
\label{fig:Tanslation prompt}
\end{figure}

\section{Detailed Results on All Model Scales and Tasks}

\label{sec:detailed_results}

\begin{figure}[t!]
    \centering
    \includegraphics[width=0.5\textwidth]{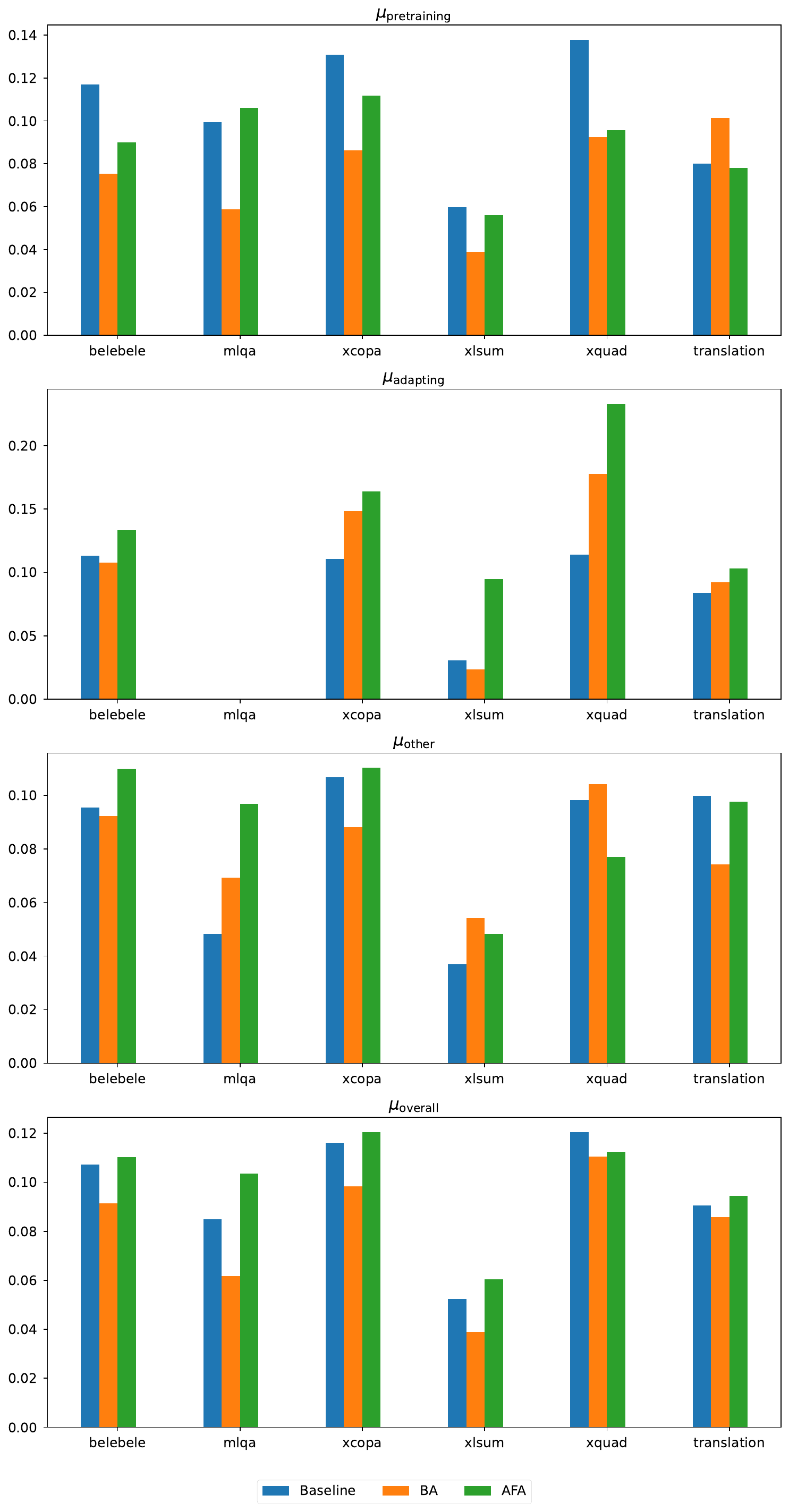}
    \caption{\textbf{Task wise performance comparisonfor 782 million parameter models.} We find that the ``Baseline Adaptation'' method is able to improve performance only on adapting languages, often at the cost of performance on all other languages.}
    \label{fig:taskwise_400m}
\end{figure}

\begin{figure}[t!]
    \centering
    \includegraphics[width=0.5\textwidth]{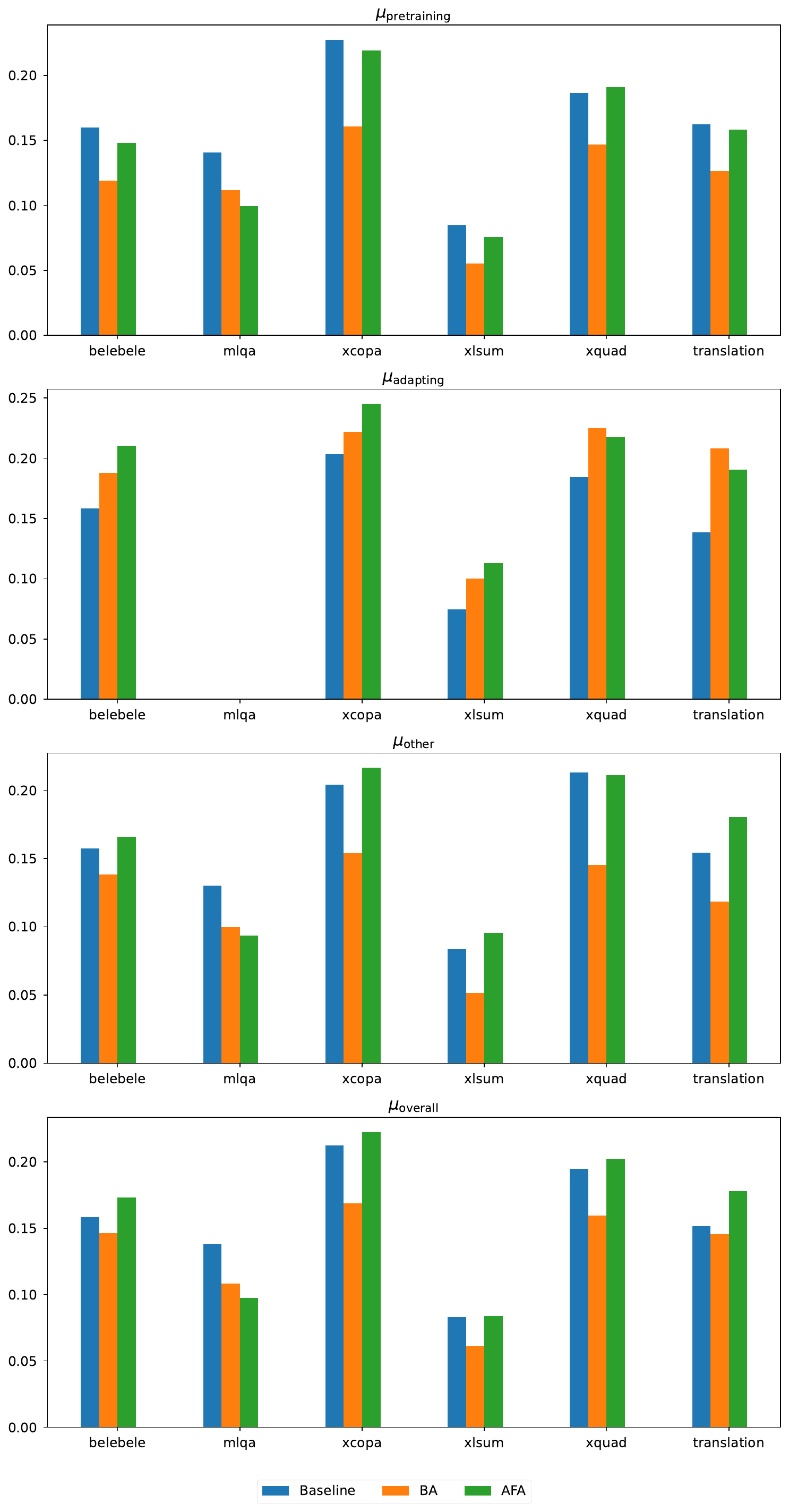}
    \caption{\textbf{Task wise performance comparison for 400 million parameter models.} We find that the ``Baseline Adaptation'' method is able to improve performance only on adapting languages, often at the cost of performance on all other languages.}
    \label{fig:taskwise_782m}
\end{figure}

Table \ref{tab:short_belebele} to Table \ref{tab:short_xquad} and Figure \ref{fig:taskwise_400m} and Figure \ref{fig:taskwise_782m} present detailed results on all tasks at all model scales and language classes.

\begin{table}
\small
\centering
\begin{tabular}{lrrrr}
\toprule
Model & $\mu_\text{pretraining}$ & $\mu_\text{adapting}$ & $\mu_\text{other}$ & $\mu_\text{overall}$ \\
\midrule
\multicolumn{5}{c}{Number of parameters = 400M}\\
\midrule
Baseline & \bfseries 0.117 & 0.113 & 0.095 & 0.107\\
BA & 0.075 & 0.108 & 0.092 & 0.091\\
AFA & 0.090 & \bfseries 0.133 & \bfseries 0.110 & \bfseries 0.110\\
\midrule
\multicolumn{5}{c}{Number of parameters = 782M}\\
\midrule
Baseline & \bfseries 0.160 & 0.158 & 0.157 & 0.158\\
BA & 0.119 & 0.188 & 0.138 & 0.146\\
AFA & 0.148 & \bfseries 0.211 & \bfseries 0.166 & \bfseries 0.173\\
\midrule
\multicolumn{5}{c}{Number of parameters = 1.6B}\\
\midrule
Baseline & \bfseries 0.184 & 0.199 & 0.198 & 0.194\\
BA & 0.167 & 0.236 & 0.156 & 0.182\\
AFA & \bfseries 0.184 & \bfseries 0.241 & \bfseries 0.206 & \bfseries 0.209\\
\midrule
\multicolumn{5}{c}{Number of parameters = 2.8B}\\
\midrule
Baseline &  0.208 & 0.224 & 0.225 & 0.219\\
BA & 0.195 & 0.261 & 0.183 & 0.208\\
AFA & \bfseries 0.215 & \bfseries 0.279 & \bfseries 0.227 & \bfseries 0.237\\
\bottomrule
\end{tabular}
\caption{Detailed results on belebele \citep{bandarkar2023belebele}. BA refers to Baseline adapted model. AFA refers to Active Forgetting adapted model. $\mu_\text{pretraining}$ refers to performance averaged over languages in the pretraining split. $\mu_\text{adapting}$ refers to averaging over languages in the adapting split. $\mu_\text{other}$ refers to averaging on languages that are in neither split. $\mu_\text{overall}$ refers to the average over all languages. We use Accuracy as the metric.}
\label{tab:short_belebele}
\end{table}
\begin{table}
\small
\centering
\begin{tabular}{lrrrr}
\toprule
Model & $\mu_\text{pretraining}$ & $\mu_\text{adapting}$ & $\mu_\text{other}$ & $\mu_\text{overall}$ \\
\midrule
\multicolumn{5}{c}{Number of parameters = 400M}\\
\midrule
Baseline & 0.099 & N/A & 0.048 & 0.085\\
BA & 0.059 & N/A & 0.069 & 0.062\\
AFA & \bfseries 0.106 & N/A & \bfseries 0.097 & \bfseries 0.104\\
\midrule
\multicolumn{5}{c}{Number of parameters = 782M}\\
\midrule
Baseline & \bfseries 0.141 & N/A & \bfseries 0.130 & \bfseries 0.138\\
BA & 0.112 & N/A & 0.100 & 0.108\\
AFA & 0.099 & N/A & 0.093 & 0.098\\
\midrule
\multicolumn{5}{c}{Number of parameters = 1.6B}\\
\midrule
Baseline & \bfseries 0.151 & N/A & \bfseries 0.248 & \bfseries 0.179\\
BA & 0.139 & N/A & 0.149 & 0.141\\
AFA & 0.135 & N/A & 0.103 & 0.126\\
\midrule
\multicolumn{5}{c}{Number of parameters = 2.8B}\\
\midrule
Baseline & \bfseries 0.183 & N/A & \bfseries 0.276 & \bfseries 0.210\\
BA & 0.171 & N/A & 0.171 & 0.171\\
AFA & 0.167 & N/A & 0.126 & 0.156\\
\bottomrule
\end{tabular}
\caption{Detailed results on mlqa \citep{lewis-etal-2020-mlqa}. BA refers to Baseline adapted model. AFA refers to Active Forgetting adapted model. $\mu_\text{pretraining}$ refers to performance averaged over languages in the pretraining split. $\mu_\text{adapting}$ refers to averaging over languages in the adapting split. $\mu_\text{other}$ refers to averaging on languages that are in neither split. $\mu_\text{overall}$ refers to the average over all languages. We use F1-abstractive score as the metric.}
\label{tab:short_mlqa}
\end{table}
\begin{table}
\small
\centering
\begin{tabular}{lrrrr}
\toprule
Model & $\mu_\text{pretraining}$ & $\mu_\text{adapting}$ & $\mu_\text{other}$ & $\mu_\text{overall}$ \\
\midrule
\multicolumn{5}{c}{Number of parameters = 400M}\\
\midrule
Baseline & \bfseries 0.131 & 0.111 & 0.107 & 0.116\\
BA & 0.086 & 0.149 & 0.088 & 0.098\\
AFA & 0.112 & \bfseries 0.164 & \bfseries 0.110 & \bfseries 0.121\\
\midrule
\multicolumn{5}{c}{Number of parameters = 782M}\\
\midrule
Baseline & \bfseries 0.227 & 0.204 & 0.204 & 0.212\\
BA & 0.161 & 0.222 & 0.154 & 0.169\\
AFA & 0.219 & \bfseries 0.245 & \bfseries 0.216 & \bfseries 0.223\\
\midrule
\multicolumn{5}{c}{Number of parameters = 1.6B}\\
\midrule
Baseline & 0.238 & 0.275 & \bfseries 0.263 & 0.256\\
BA & 0.200 & 0.319 & 0.210 & 0.226\\
AFA & \bfseries 0.254 & \bfseries 0.321 & \bfseries 0.263 & \bfseries 0.270\\
\midrule
\multicolumn{5}{c}{Number of parameters = 2.8B}\\
\midrule
Baseline & 0.278 & 0.319 & 0.288 & 0.295\\
BA & 0.243 & \bfseries 0.350 & 0.253 & 0.282\\
AFA & \bfseries 0.279 &  0.346 & \bfseries 0.299 & \bfseries 0.308\\
\bottomrule
\end{tabular}
\caption{Detailed results on xcopa \citep{ponti-etal-2020-xcopa}. BA refers to Baseline adapted model. AFA refers to Active Forgetting adapted model. $\mu_\text{pretraining}$ refers to performance averaged over languages in the pretraining split. $\mu_\text{adapting}$ refers to averaging over languages in the adapting split. $\mu_\text{other}$ refers to averaging on languages that are in neither split. $\mu_\text{overall}$ refers to the average over all languages. We use accuracy as the metric.}
\label{tab:short_xcopa}
\end{table}
\begin{table}
\small
\centering
\begin{tabular}{lrrrr}
\toprule
Model & $\mu_\text{pretraining}$ & $\mu_\text{adapting}$ & $\mu_\text{other}$ & $\mu_\text{overall}$ \\
\midrule
\multicolumn{5}{c}{Number of parameters = 400M}\\
\midrule
Baseline & \bfseries 0.060 & 0.031 & 0.037 & 0.052\\
BA & 0.039 & 0.024 & \bfseries 0.054 & 0.039\\
AFA & 0.056 & \bfseries 0.095 & 0.048 & \bfseries 0.060\\
\midrule
\multicolumn{5}{c}{Number of parameters = 782M}\\
\midrule
Baseline & \bfseries 0.085 & 0.074 & 0.084 & 0.083\\
BA & 0.055 & 0.100 & 0.052 & 0.061\\
AFA & 0.076 & \bfseries 0.113 & \bfseries 0.095 & \bfseries 0.084\\
\midrule
\multicolumn{5}{c}{Number of parameters = 1.6B}\\
\midrule
Baseline & 0.096 & 0.119 & \bfseries 0.110 & 0.101\\
BA & 0.076 & 0.115 & 0.058 & 0.079\\
AFA & \bfseries 0.097 & \bfseries 0.134 & 0.109 & \bfseries 0.104\\
\midrule
\multicolumn{5}{c}{Number of parameters = 2.8B}\\
\midrule
Baseline & \bfseries 0.122 & 0.141 & \bfseries 0.138 & \bfseries 0.127\\
BA & 0.101 & 0.119 & 0.082 & 0.100\\
AFA &  0.110 & \bfseries 0.162 & 0.114 &  0.124\\
\bottomrule
\end{tabular}
\caption{Detailed results on xlsum \citep{hasan-etal-2021-xl}. BA refers to Baseline adapted model. AFA refers to Active Forgetting adapted model. $\mu_\text{pretraining}$ refers to performance averaged over languages in the pretraining split. $\mu_\text{adapting}$ refers to averaging over languages in the adapting split. $\mu_\text{other}$ refers to averaging on languages that are in neither split. $\mu_\text{overall}$ refers to the average over all languages. We use Rouge Score as the metric.}
\label{tab:short_xlsum}
\end{table}
\begin{table}
\small
\centering
\begin{tabular}{lrrrr}
\toprule
Model & $\mu_\text{pretraining}$ & $\mu_\text{adapting}$ & $\mu_\text{other}$ & $\mu_\text{overall}$ \\
\midrule
\multicolumn{5}{c}{Number of parameters = 400M}\\
\midrule
Baseline & \bfseries 0.138 & 0.114 & 0.098 & \bfseries 0.121\\
BA & 0.093 & 0.178 & \bfseries 0.104 & 0.111\\
AFA & 0.096 & \bfseries 0.233 & 0.077 & 0.112\\
\midrule
\multicolumn{5}{c}{Number of parameters = 782M}\\
\midrule
Baseline & 0.186 & 0.185 & \bfseries 0.213 & 0.195\\
BA & 0.147 & \bfseries 0.225 & 0.145 & 0.159\\
AFA & \bfseries 0.191 & 0.218 & 0.211 & \bfseries 0.202\\
\midrule
\multicolumn{5}{c}{Number of parameters = 1.6B}\\
\midrule
Baseline & \bfseries 0.243 & 0.248 & 0.231 & 0.240\\
BA & 0.195 & 0.285 & 0.216 & 0.217\\
AFA & 0.242 & \bfseries 0.335 & \bfseries 0.258 & \bfseries 0.263\\
\midrule
\multicolumn{5}{c}{Number of parameters = 2.8B}\\
\midrule
Baseline &  0.258 & 0.277 & 0.254 & 0.263\\
BA & 0.230 & 0.320 & 0.230 & 0.260\\
AFA & \bfseries 0.265 & \bfseries 0.361 & \bfseries 0.230 & \bfseries 0.308\\
\bottomrule
\end{tabular}
\caption{Detailed results on xquad \citep{artetxe-etal-2020-cross}. BA refers to Baseline adapted model. AFA refers to Active Forgetting adapted model. $\mu_\text{pretraining}$ refers to performance averaged over languages in the pretraining split. $\mu_\text{adapting}$ refers to averaging over languages in the adapting split. $\mu_\text{other}$ refers to averaging on languages that are in neither split. $\mu_\text{overall}$ refers to the average over all languages. We use F1 abstractive as the metric.}
\label{tab:short_xquad}
\end{table}

\begin{figure}[t!]
    \centering
    \includegraphics[width=0.5\textwidth]{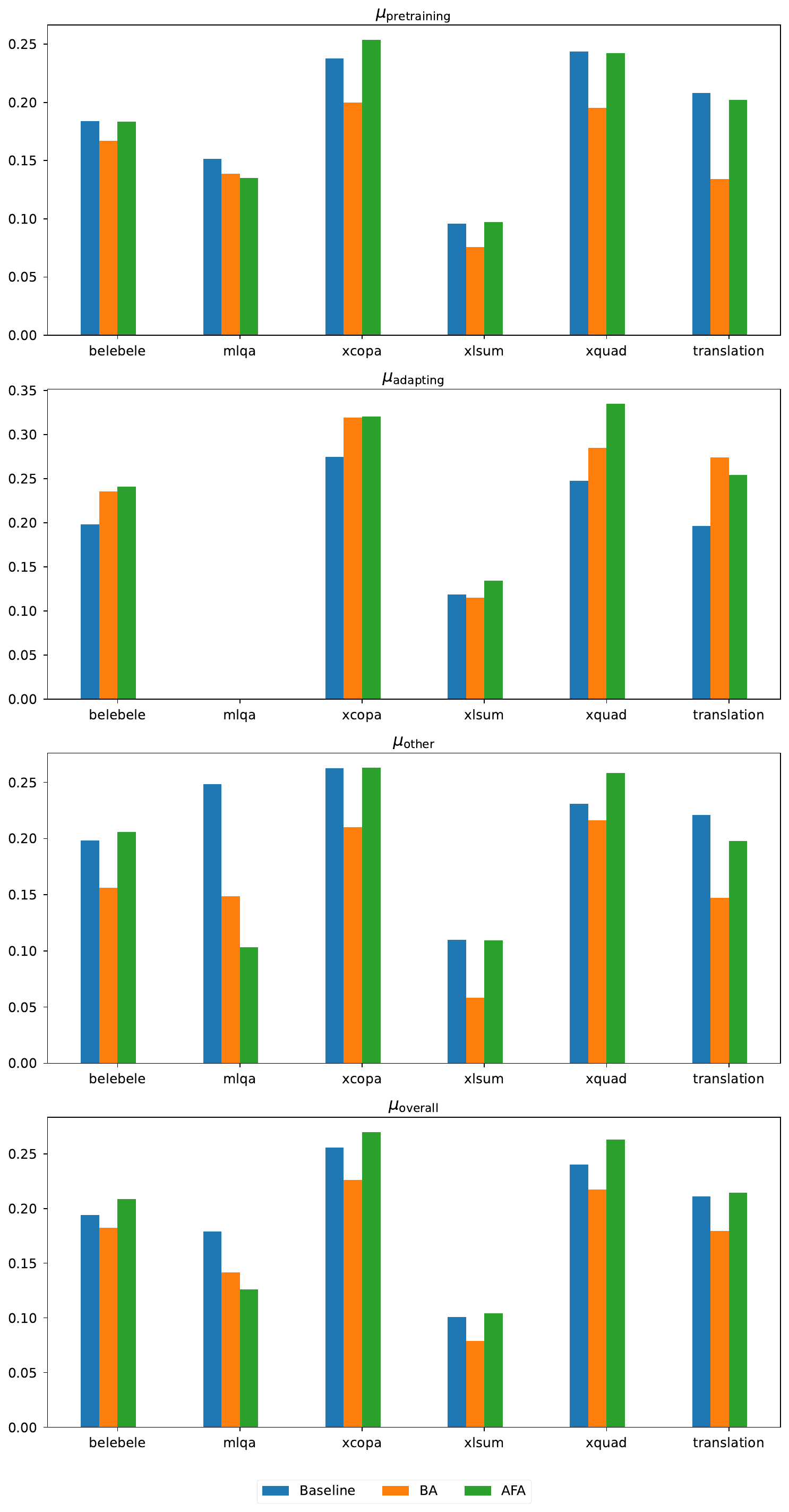}
    \caption{\textbf{Task wise performance comparison for the 1.6 billion parameter models.}
    We find that the ``Baseline Adaptation'' method is able to improve performance only on adapting languages, often at the cost of performance on all other languages. 
    }
    \label{fig:taskwise_1.6b}
\end{figure}

\end{document}